\documentclass[11pt]{article}

\usepackage[utf8]{inputenc}
\usepackage[T1]{fontenc}
\usepackage{times}
\usepackage{latexsym}
\usepackage{amsmath,amssymb,amsfonts}
\usepackage{graphicx}
\usepackage{booktabs}
\usepackage{multirow}
\usepackage{xcolor}
\usepackage{hyperref}
\usepackage{caption}
\usepackage{subcaption}
\usepackage{enumitem}
\usepackage{natbib}
\usepackage[margin=1in]{geometry}
\usepackage{tikz}
\usepackage{pgfplots}
\pgfplotsset{compat=1.17}
\usetikzlibrary{positioning, arrows.meta, shapes.geometric}

\hypersetup{
    colorlinks=true,
    linkcolor=blue,
    citecolor=blue,
    urlcolor=blue
}

\definecolor{deepseekblue}{RGB}{0,114,189}
\definecolor{gptgreen}{RGB}{76,175,80}
\definecolor{claudepurple}{RGB}{156,39,176}
\definecolor{detectblue}{RGB}{33,150,243}
\definecolor{localorange}{RGB}{255,152,0}
\definecolor{correctgreen}{RGB}{76,175,80}

\title{Decomposing LLM Self-Correction: \\
\large The Accuracy-Correction Paradox and Error Depth Hypothesis}

\author{
  Yin Li \\
  University of Birmingham \\
  \texttt{kxl474@student.bham.ac.uk} \\
}

\usepackage{float}
\begin{document}

\maketitle

\begin{abstract}
Large Language Models (LLMs) are widely believed to possess self-correction capabilities, yet recent studies suggest that \textit{intrinsic} self-correction—where models correct their own outputs without external feedback—remains largely ineffective. In this work, we systematically decompose self-correction into three distinct sub-capabilities: \textbf{error detection}, \textbf{error localization}, and \textbf{error correction}. Through cross-model experiments on GSM8K-Complex (n=500 per model, 346 total errors) with three major LLMs, we uncover a striking \textbf{Accuracy-Correction Paradox}: \textit{weaker} models (GPT-3.5, 66\% accuracy) achieve 1.6$\times$ \textit{higher} intrinsic correction rates than stronger models (DeepSeek, 94\% accuracy)—26.8\% vs 16.7\%. We propose the \textbf{Error Depth Hypothesis}: stronger models make fewer but ``deeper'' errors that resist self-correction. Error detection rates vary dramatically across architectures (10\% to 82\%), yet detection capability does not predict correction success—Claude detects only 10\% of errors but corrects 29\% intrinsically. Surprisingly, providing error location hints \textit{hurts} all models. Our findings challenge linear assumptions about model capability and self-improvement, with important implications for the design of self-refinement pipelines.
\end{abstract}


\section{Introduction}

The ability to recognize and correct one's own mistakes is a hallmark of intelligent reasoning. As Large Language Models (LLMs) are increasingly deployed in high-stakes applications—from mathematical problem-solving to code generation and scientific reasoning—understanding their capacity for \textit{self-correction} has become critically important.

Recent work has investigated whether LLMs can improve their outputs through self-refinement \citep{madaan2023selfrefine, shinn2023reflexion}. While some studies report improvements through iterative prompting, a growing body of evidence suggests that \textit{intrinsic} self-correction—where models correct errors without external validation signals—is fundamentally limited \citep{huang2024large}.

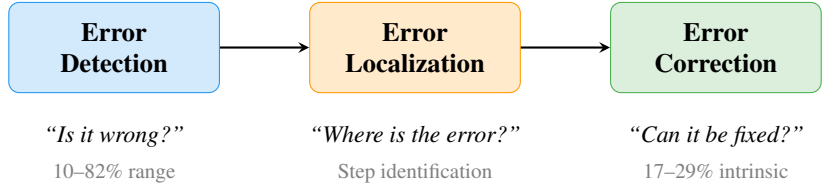
\begin{figure}[h]
\centering
\begin{tikzpicture}[
    box/.style={rectangle, draw, rounded corners, minimum width=2.8cm, minimum height=1.2cm, align=center, font=\small\bfseries},
    arrow/.style={->, thick, >=stealth}
]
\node[box, fill=detectblue!20, draw=detectblue] (detect) at (0,0) {Error\\Detection};
\node[box, fill=localorange!20, draw=localorange] (local) at (4,0) {Error\\Localization};
\node[box, fill=correctgreen!20, draw=correctgreen] (correct) at (8,0) {Error\\Correction};

\draw[arrow] (detect) -- (local);
\draw[arrow] (local) -- (correct);

\node[below=0.3cm of detect, font=\footnotesize\itshape] {``Is it wrong?''};
\node[below=0.3cm of local, font=\footnotesize\itshape] {``Where is the error?''};
\node[below=0.3cm of correct, font=\footnotesize\itshape] {``Can it be fixed?''};

\node[below=0.8cm of detect, font=\scriptsize, text=gray] {10--82\% range};
\node[below=0.8cm of local, font=\scriptsize, text=gray] {Step identification};
\node[below=0.8cm of correct, font=\scriptsize, text=gray] {17--29\% intrinsic};
\end{tikzpicture}
\caption{The Self-Correction Decomposition Framework. We separate self-correction into three distinct, independently measurable capabilities.}
\label{fig:framework}
\end{figure}

In this paper, we argue that the concept of ``self-correction'' conflates several distinct capabilities that deserve independent study. We propose a \textbf{decomposition framework} that separates self-correction into three measurable sub-capabilities (Figure~\ref{fig:framework}):

\begin{enumerate}[leftmargin=*]
    \item \textbf{Error Detection}: Can the model identify that its output contains an error?
    \item \textbf{Error Localization}: Can it pinpoint \textit{where} the error occurs?
    \item \textbf{Error Correction}: Can it produce a corrected solution?
\end{enumerate}

By decomposing self-correction into these sub-capabilities, we reveal important and \textit{counterintuitive} insights about the limitations of current LLMs. Our key contributions are:

\begin{itemize}[leftmargin=*]
    \item We propose a novel \textbf{decomposition framework} for analyzing LLM self-correction, separating detection, localization, and correction into independently measurable capabilities.
    
    \item We discover the \textbf{Accuracy-Correction Paradox}: weaker models (GPT-3.5, Claude) achieve 1.6--1.7$\times$ higher intrinsic correction rates than stronger models (DeepSeek).
    
    \item We propose the \textbf{Error Depth Hypothesis}: stronger models make ``deeper'' errors that resist self-correction, while weaker models make ``shallower'' errors that are easily fixable.
    
    \item We demonstrate that \textbf{error detection is architecture-dependent}, with Claude achieving only 10\% detection vs 82\% for GPT-3.5—an 8$\times$ difference.
\end{itemize}

\section{Related Work}

\paragraph{Self-Refinement in LLMs.}
Self-Refine \citep{madaan2023selfrefine} proposed iterative refinement where models generate, critique, and revise their outputs. Reflexion \citep{shinn2023reflexion} incorporated verbal reinforcement learning for multi-step tasks. However, \citet{huang2024large} demonstrated that without external oracles, LLMs cannot reliably self-correct reasoning errors—a finding our work extends with cross-model analysis.

\paragraph{Error Detection and Verification.}
\citet{lightman2023lets} explored process reward models for verifying intermediate reasoning steps. \citet{cobbe2021training} introduced verifiers for mathematical problem-solving. Our work complements this by examining whether models can verify \textit{their own} outputs and how this varies across architectures.

\paragraph{Mathematical Reasoning.}
Chain-of-thought prompting \citep{wei2022chain} and related techniques have substantially improved LLM performance on mathematical reasoning. GSM8K \citep{cobbe2021gsm8k} has become a standard benchmark for evaluating these capabilities.

\section{Methodology}

\subsection{Problem Formulation}

Given a mathematical reasoning problem $P$, an LLM generates a solution $S = \{s_1, s_2, \ldots, s_n\}$ consisting of $n$ reasoning steps, with a final answer $A$. If $A \neq A^*$ (the gold answer), we say $S$ contains an error. We decompose self-correction into three tasks:

\begin{enumerate}[leftmargin=*]
    \item \textbf{Error Detection}: Given $(P, S)$, predict whether $S$ is correct or incorrect.
    \item \textbf{Error Localization}: Given $(P, S)$ where $S$ is incorrect, identify the step $k$ where the first error occurs.
    \item \textbf{Error Correction}: Given $(P, S)$ and optionally the error location $k$, produce a corrected solution $S'$ with $A' = A^*$.
\end{enumerate}

\subsection{Experimental Setup}

\paragraph{Dataset.} We use GSM8K \citep{cobbe2021gsm8k}, a dataset of grade-school math word problems requiring multi-step arithmetic reasoning. We sample problems using a fixed random seed (42) for reproducibility.

\paragraph{GSM8K-Complex (Ours).} To rigorously test self-correction on capable models, we introduce \textbf{GSM8K-Complex}, a subset of 500 problems filtered for higher complexity. We select problems meeting \textit{at least 2 of 3} criteria: (1) question length $>$ 100 characters; (2) solution contains $\ge$ 4 computation steps (counted by ``<<'' markers); (3) solution contains $\ge$ 3 distinct operations. Problem indices will be released with code.

\paragraph{Models.} We evaluate three models representing different capability levels and provider architectures:
\begin{itemize}[leftmargin=*]
    \item \textbf{DeepSeek-Chat}: A capable instruction-tuned model (94\% baseline accuracy on GSM8K-Complex)
    \item \textbf{GPT-3.5-Turbo}: OpenAI's efficient model (68\% baseline accuracy)
    \item \textbf{Claude-3-Haiku}: Anthropic's fast model (73.3\% baseline accuracy)
\end{itemize}

\paragraph{Evaluation Protocol.} For each model, we:
\begin{enumerate}[leftmargin=*]
    \item Generate solutions for all problems
    \item Collect incorrect solutions (where model answer $\neq$ gold answer)
    \item Test \textbf{Error Detection}: Does the model correctly classify its solution as incorrect?
    \item Test \textbf{Correction with Hint}: Given the error step location, can the model correct the solution?
    \item Test \textbf{Intrinsic Correction}: Without hints, can the model correct the solution?
    \item Test \textbf{Iterative Reflection}: Over up to 3 rounds, can verification and re-solving succeed?
\end{enumerate}

\subsection{Metrics}

\begin{itemize}[leftmargin=*]
    \item \textbf{Detection Accuracy}: Fraction of incorrect solutions correctly identified as incorrect
    \item \textbf{Correction Success Rate}: Fraction of incorrect solutions successfully corrected to the gold answer
    \item \textbf{Iterative Success Rate}: Maximum correction rate achieved across multiple reflection rounds
\end{itemize}

\section{Results}

\subsection{Cross-Model Comparison}

Table~\ref{tab:main_results} presents the main findings across all three models.

\begin{table}[h]
\centering
\caption{Cross-Model Self-Correction Performance on GSM8K-Complex (n=500 per model)}
\label{tab:main_results}
\small
\begin{tabular}{lcccccc}
\toprule
\textbf{Model} & \textbf{Acc.} & \textbf{Errors} & \textbf{Detection} & \textbf{Intrinsic} & \textbf{With Hint} & \textbf{Iterative} \\
\midrule
DeepSeek & 94.0\% & 30 & 17/30 (56.7\%) & 5/30 (16.7\%) & 8/30 (26.7\%) & 6/30 (20.0\%) \\
 & & & {\scriptsize [37, 75]} & {\scriptsize [6, 35]} & {\scriptsize [12, 46]} & {\scriptsize [8, 39]} \\
\midrule
GPT-3.5 & 66.4\% & 168 & 137/168 (81.5\%) & 45/168 (26.8\%) & 26/168 (15.5\%) & 114/168 (67.9\%) \\
 & & & {\scriptsize [75, 87]} & {\scriptsize [20, 34]} & {\scriptsize [10, 22]} & {\scriptsize [60, 75]} \\
\midrule
Claude & 70.4\% & 148 & 15/148 (10.1\%) & 43/148 (29.1\%) & 19/148 (12.8\%) & 90/148 (60.8\%) \\
 & & & {\scriptsize [6, 16]} & {\scriptsize [22, 37]} & {\scriptsize [8, 19]} & {\scriptsize [53, 69]} \\
\bottomrule
\end{tabular}
\vspace{0.1cm}

\small{\textit{Note}: 95\% Clopper-Pearson confidence intervals shown in brackets. \textit{Detection}: correctly identifying error exists. \textit{Intrinsic}: correction without hints. \textit{Iterative}: up to 3 reflection rounds.}
\end{table}

\subsection{The Accuracy-Correction Paradox}

\begin{figure}[h]
\centering
\begin{tikzpicture}
\begin{axis}[
    ybar,
    bar width=0.4cm,
    width=0.9\textwidth,
    height=6cm,
    ylabel={Percentage (\%)},
    symbolic x coords={DeepSeek, GPT-3.5, Claude},
    xtick=data,
    ymin=0, ymax=100,
    legend style={at={(0.5,1.15)}, anchor=south, legend columns=2},
    enlarge x limits=0.25,
    nodes near coords,
    every node near coord/.append style={font=\scriptsize},
]
\addplot[fill=deepseekblue!70] coordinates {(DeepSeek, 94) (GPT-3.5, 66.4) (Claude, 70.4)};
\addplot[fill=correctgreen!70] coordinates {(DeepSeek, 16.7) (GPT-3.5, 26.8) (Claude, 29.1)};
\legend{Model Accuracy, Intrinsic Correction}
\end{axis}
\end{tikzpicture}
\caption{The Accuracy-Correction Paradox. The \textit{strongest} model (DeepSeek, 94\% accuracy) achieves the \textit{lowest} intrinsic correction rate (16.7\%), while weaker models correct 1.6--1.7$\times$ more errors.}
\label{fig:paradox}
\end{figure}
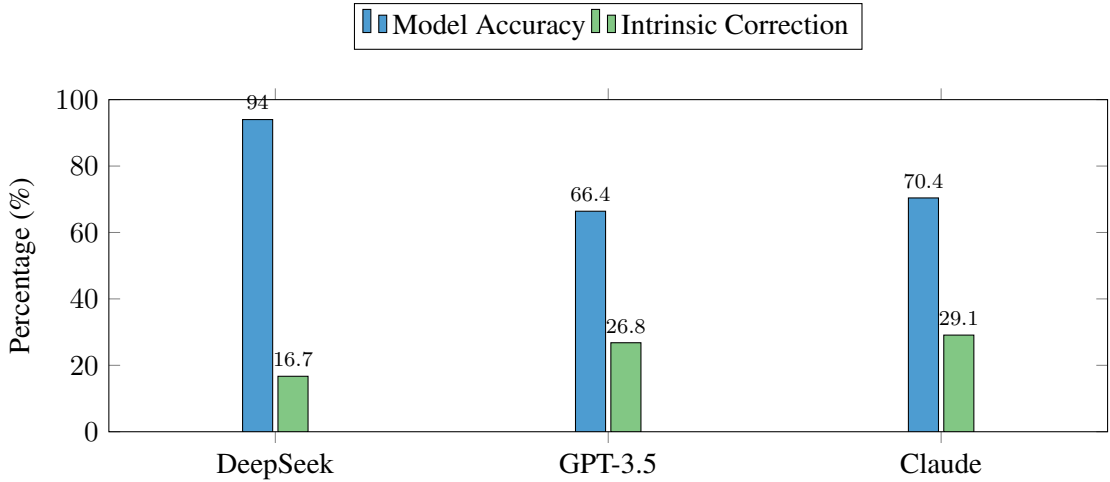

Figure~\ref{fig:paradox} visualizes our key finding: the inverse relationship between model accuracy and intrinsic correction capability.

\paragraph{Finding 1: The Accuracy-Correction Paradox.}
The strongest model (DeepSeek, 94\% accuracy) achieves the \textbf{lowest} intrinsic correction rate (16.7\%), while weaker models (GPT-3.5: 26.8\%, Claude: 29.1\%) correct 1.6--1.7$\times$ more errors. This suggests that model capability does not linearly translate to self-correction ability.

\paragraph{Finding 2: Detection Does Not Predict Correction.}
GPT-3.5 detects 81.5\% of errors but corrects only 26.8\%. Claude detects only 10.1\% but corrects 29.1\%—achieving \textbf{higher} correction with 8$\times$ \textit{worse} detection. Detection and correction are largely independent capabilities.

\paragraph{Finding 3: Model-Generated Hints Hurt.}
Surprisingly, providing error location hints from the model's own localization \textit{decreases} correction rates for all models (GPT-3.5: 26.8\%$\rightarrow$15.5\%, Claude: 29.1\%$\rightarrow$12.8\%). Since hints are model-generated (not ground-truth), this may reflect poor localization quality or anchoring to incorrect reasoning paths.

\paragraph{Finding 4: Iterative Reflection Compensates for Weak Detection.}
Despite Claude's 10\% detection rate, iterative reflection achieves 60.8\% correction—a 6$\times$ improvement. Multi-round prompting bypasses detection limitations through repeated re-solving.

\subsection{Error Type Analysis}

Table~\ref{tab:error_types} reveals the distribution of error types, supporting our Error Depth Hypothesis.

\begin{table}[h]
\centering
\caption{Error Type Distribution Across Models}
\label{tab:error_types}
\begin{tabular}{lccc}
\toprule
\textbf{Error Type} & \textbf{DeepSeek} & \textbf{GPT-3.5} & \textbf{Claude} \\
\midrule
Setup/Interpretation & 44\% & 25\% & 38\% \\
Logic Error & 33\% & 13\% & 25\% \\
Calculation Error & 22\% & \textbf{62\%} & 37\% \\
\bottomrule
\end{tabular}
\end{table}

\paragraph{Labeling Methodology.} Error types are assigned automatically using the model's own error localization capability (see Appendix A.2 for prompt). Each model classifies its errors into: CALCULATION (arithmetic mistakes), LOGIC (incorrect reasoning), or SETUP (problem misinterpretation). This approach provides scalable labeling but may underestimate certain error types if the model lacks self-awareness. We acknowledge this as a limitation; future work should validate against human annotations.

A crucial pattern emerges: GPT-3.5's errors are predominantly \textit{calculation errors} (62\%), which are typically ``shallower'' and easier to self-correct. DeepSeek's errors are mostly \textit{setup and logic errors} (77\%), representing deeper reasoning failures that resist intrinsic correction.

\subsection{Iterative Reflection Dynamics}

\begin{figure}[h]
\centering
\begin{tikzpicture}
\begin{axis}[
    width=0.85\textwidth,
    height=5.5cm,
    xlabel={Reflection Round},
    ylabel={Cumulative Correction Rate (\%)},
    xmin=0.5, xmax=3.5,
    ymin=0, ymax=80,
    xtick={1,2,3},
    legend style={at={(0.02,0.98)}, anchor=north west},
    grid=major,
    mark size=3pt,
]
\addplot[color=deepseekblue, mark=square*, thick] coordinates {(1, 16.7) (2, 20.0) (3, 20.0)};
\addplot[color=gptgreen, mark=*, thick] coordinates {(1, 26.8) (2, 50.6) (3, 67.9)};
\addplot[color=claudepurple, mark=triangle*, thick] coordinates {(1, 29.1) (2, 47.3) (3, 60.8)};
\legend{DeepSeek, GPT-3.5, Claude}
\end{axis}
\end{tikzpicture}
\caption{Iterative Reflection Dynamics. DeepSeek saturates after round 1 (20\%), while GPT-3.5 and Claude continue improving through round 3 (68\%, 61\%).}
\label{fig:iterative}
\end{figure}
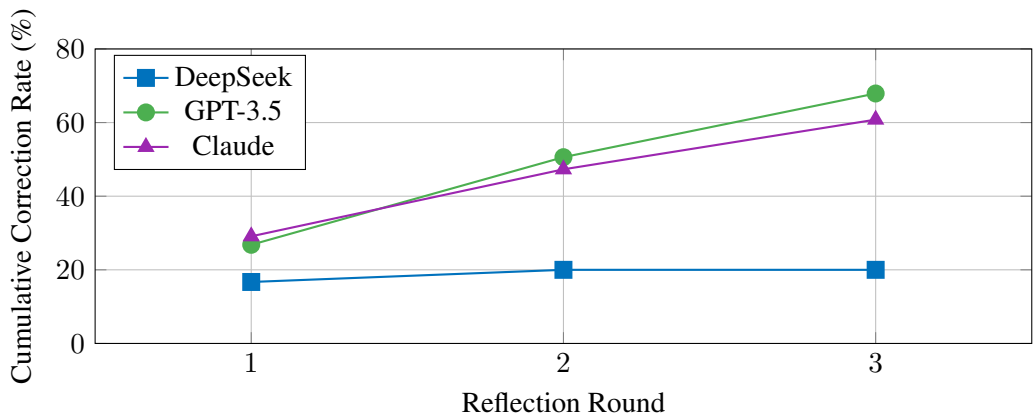

Figure~\ref{fig:iterative} shows how correction rates evolve across reflection rounds.

DeepSeek saturates after round 1 with minimal gains (20\%), while GPT-3.5 and Claude continue improving through round 3 (68\%, 61\%). This confirms that models with ``shallower'' errors benefit more from iterative approaches.

\section{Analysis and Discussion}

\subsection{The Error Depth Hypothesis}

Why does a weaker model correct better? We propose the \textbf{Error Depth Hypothesis}:

\begin{quote}
\textit{Stronger models make fewer but ``deeper'' errors (setup, logic) that are fundamentally harder to self-correct. Weaker models make more but ``shallower'' errors (calculation) that are easily fixable upon re-examination.}
\end{quote}

This hypothesis is supported by our error type analysis (Table~\ref{tab:error_types}): 62\% of GPT-3.5 errors are calculation mistakes, vs only 22\% for DeepSeek. Calculation errors often involve simple arithmetic that the model can correct when prompted to ``check carefully.'' Setup and logic errors require fundamental re-thinking of the problem approach—something models struggle to do without external guidance.

\subsection{Architecture-Dependent Detection}

The dramatic variation in detection rates (10\% to 82\%) across architectures suggests that self-verification is not a universal capability. Claude's low detection rate despite reasonable correction via iteration indicates that some models may ``accidentally'' correct errors through re-solving without explicitly recognizing them.

\subsection{Implications for Self-Refinement Systems}

Our findings have important practical implications:

\begin{enumerate}[leftmargin=*]
    \item \textbf{Self-correction efficacy depends on error types.} Systems should characterize the error distribution before deploying self-correction strategies.
    
    \item \textbf{Stronger models may need different interventions.} Model-generated hints hurt all models in our study, suggesting localization-based feedback requires higher quality hints (e.g., human-annotated).
    
    \item \textbf{Iterative reflection is more valuable for weaker models.} Multi-round prompting yields 3$\times$ improvement for GPT-3.5/Claude but minimal gains for DeepSeek.
    
    \item \textbf{Detection $\neq$ Correction across all models.} Claude detects 10\% but corrects 29\%; GPT-3.5 detects 82\% but corrects 27\%.
\end{enumerate}

\subsection{Limitations}

\paragraph{Sample Size.} While GSM8K-Complex yields 346 total errors (DeepSeek: 30, GPT-3.5: 168, Claude: 148), DeepSeek's high accuracy still limits statistical power for that model. Future work should use even harder benchmarks or synthetic error injection.

\paragraph{Three Models.} Extending to additional models (GPT-4, Claude-Sonnet, Llama-3) would further validate the accuracy-correction paradox.

\paragraph{Hint Oracle.} Our ``with hint'' condition uses model-generated step localization, not ground-truth annotations. The surprising negative effect of hints warrants investigation with human-annotated error locations.

\paragraph{Domain Specificity.} Mathematical reasoning has ground truth; findings may differ for open-ended generation tasks.

\section{Conclusion}

We have presented a decomposition framework for analyzing LLM self-correction, separating error detection, localization, and correction as distinct capabilities. Our experiments on GSM8K-Complex (n=500 per model, 346 total errors) reveal the striking \textbf{Accuracy-Correction Paradox}: the strongest model (DeepSeek) achieves the lowest intrinsic correction rate (17\%), while weaker models correct 1.6--1.7$\times$ more errors. We propose the \textbf{Error Depth Hypothesis}: stronger models make ``deeper'' errors that resist self-correction.

Surprisingly, providing error location hints \textit{hurts} all models, while iterative reflection compensates dramatically for weak detection (Claude: 10\% detect $\rightarrow$ 61\% iterative). Our findings challenge linear assumptions about model capability and self-improvement, with important implications for the design of self-refinement pipelines.

\section{Future Work}

Future directions include: (1) scaling experiments to larger sample sizes across additional models including GPT-4, Claude-Sonnet, and Llama-3; (2) investigating whether fine-tuning on self-correction data improves intrinsic correction rates; (3) developing hybrid systems that combine model-based detection with tool-augmented correction (e.g., calculators); and (4) extending the analysis to other reasoning domains including code generation and commonsense reasoning.

\section*{Reproducibility Statement}

All experiments use publicly available models accessed via API. We provide complete prompt templates and evaluation scripts. Random seed is fixed at 42 for dataset sampling. Full experiment code and results are available at: \url{https://github.com/Kevin0304-li/llm-self-correction}.

\bibliographystyle{plainnat}

\begin{thebibliography}{99}

\bibitem[Cobbe et al.(2021)]{cobbe2021gsm8k}
Karl Cobbe, Vineet Kosaraju, Mohammad Bavarian, Mark Chen, Heewoo Jun, Lukasz Kaiser, Matthias Plappert, Jerry Tworek, Jacob Hilton, Reiichiro Nakano, et al.
\newblock Training verifiers to solve math word problems.
\newblock \textit{arXiv preprint arXiv:2110.14168}, 2021.

\bibitem[Cobbe et al.(2021)]{cobbe2021training}
Karl Cobbe, Vineet Kosaraju, Mohammad Bavarian, Mark Chen, Heewoo Jun, Lukasz Kaiser, Matthias Plappert, Jerry Tworek, Jacob Hilton, Reiichiro Nakano, et al.
\newblock Training verifiers to solve math word problems.
\newblock \textit{arXiv preprint arXiv:2110.14168}, 2021.

\bibitem[Huang et al.(2024)]{huang2024large}
Jie Huang, Xinyun Chen, Swaroop Mishra, Huaixiu Steven Zheng, Adams Wei Yu, Xinying Song, and Denny Zhou.
\newblock Large language models cannot self-correct reasoning yet.
\newblock In \textit{International Conference on Learning Representations (ICLR)}, 2024.

\bibitem[Lightman et al.(2023)]{lightman2023lets}
Hunter Lightman, Vineet Kosaraju, Yura Burda, Harri Edwards, Bowen Baker, Teddy Lee, Jan Leike, John Schulman, Ilya Sutskever, and Karl Cobbe.
\newblock Let's verify step by step.
\newblock \textit{arXiv preprint arXiv:2305.20050}, 2023.

\bibitem[Madaan et al.(2023)]{madaan2023selfrefine}
Aman Madaan, Niket Tandon, Prakhar Gupta, Skyler Hallinan, Luyu Gao, Sarah Wiegreffe, Uri Alon, Nouha Dziri, Shrimai Prabhumoye, Yiming Yang, et al.
\newblock Self-refine: Iterative refinement with self-feedback.
\newblock In \textit{Advances in Neural Information Processing Systems (NeurIPS)}, 2023.

\bibitem[Shinn et al.(2023)]{shinn2023reflexion}
Noah Shinn, Federico Cassano, Ashwin Gopinath, Karthik R Narasimhan, and Shunyu Yao.
\newblock Reflexion: Language agents with verbal reinforcement learning.
\newblock In \textit{Advances in Neural Information Processing Systems (NeurIPS)}, 2023.

\bibitem[Wei et al.(2022)]{wei2022chain}
Jason Wei, Xuezhi Wang, Dale Schuurmans, Maarten Bosma, Fei Xia, Ed Chi, Quoc V Le, Denny Zhou, et al.
\newblock Chain-of-thought prompting elicits reasoning in large language models.
\newblock In \textit{Advances in Neural Information Processing Systems (NeurIPS)}, 2022.

\end{thebibliography}

\appendix
\section{Prompt Templates}

\subsection{Error Detection Prompt}
\begin{verbatim}
Look at this solution to a math problem. Is the solution correct?

Question: {question}
Solution: {solution}
Final Answer Given: {predicted_answer}

Analyze the solution carefully. Is there any error?

Respond with:
VERDICT: CORRECT or INCORRECT
CONFIDENCE: HIGH, MEDIUM, or LOW
EXPLANATION: (brief explanation)
\end{verbatim}

\subsection{Intrinsic Correction Prompt}
\begin{verbatim}
Please verify your previous solution and correct any errors.

Question: {question}
Your previous solution: {solution}
Your previous answer: {predicted_answer}

Please carefully check each step. If you find any errors, 
provide the corrected solution and final answer.
\end{verbatim}

\section{Detailed Results}

\begin{table}[h]
\centering
\caption{Detailed Experimental Configuration}
\begin{tabular}{lc}
\toprule
\textbf{Parameter} & \textbf{Value} \\
\midrule
Dataset & GSM8K (test split) \\
Random Seed & 42 \\
Temperature & 0.0 \\
Max Tokens & 2048 \\
Iterative Rounds & 3 \\
\bottomrule
\end{tabular}
\end{table}

\section{GSM8K-Complex Problem IDs}

The GSM8K-Complex (Ours) subset consists of 500 problems filtered from the GSM8K test set. We select problems meeting \textit{at least 2 of 3} criteria: (1) question length $>$ 100 characters; (2) solution contains $\ge$ 4 computation steps (counted by ``<<'' markers); (3) solution contains $\ge$ 3 distinct mathematical operations. The complete list of problem indices (0-indexed from the GSM8K test split) will be released with our code repository.

\end{document}